\documentclass[conference]{IEEEtran}
\IEEEoverridecommandlockouts
\usepackage{cite}
\usepackage{amsmath,amssymb,amsfonts}
\usepackage{algorithmic}
\usepackage{graphicx}
\usepackage{textcomp}
\usepackage{xcolor}
\usepackage{color}
\usepackage{tabularray}
\usepackage{threeparttable}
\usepackage{multirow}
\usepackage{booktabs}
\usepackage[most]{tcolorbox}
\def\BibTeX{{\rm B\kern-.05em{\sc i\kern-.025em b}\kern-.08em
    T\kern-.1667em\lower.7ex\hbox{E}\kern-.125emX}}
\begin{document}

\title{Exploring the Comprehension of ChatGPT in Traditional Chinese Medicine Knowledge \\



}

\author{\IEEEauthorblockN{1\textsuperscript{st} Yizhen Li}
\IEEEauthorblockA{\textit{School of Computer Science and Engineer} \\
\textit{Beihang University}\\
Beijing, China \\
liyizhen@buaa.edu.cn}
\and
\IEEEauthorblockN{2\textsuperscript{nd} Shaohan Huang}
\IEEEauthorblockA{\textit{School of Computer Science and Engineer} \\
\textit{Beihang University}\\
Beijing, China \\
huangshaohan@buaa.edu.cn}
\and
\IEEEauthorblockN{3\textsuperscript{rd} Jiaxing Qi}
\IEEEauthorblockA{\textit{School of Computer Science and Engineer} \\
\textit{Beihang University}\\
Beijing, China \\
jiaxingqi@buaa.edu.cn}
\and
\IEEEauthorblockN{4\textsuperscript{th} Lei Quan}
\IEEEauthorblockA{\textit{School of Life and Science} \\
\textit{Beijing University of Chinese Medicine}\\
Beijing, China \\
tristone932473910@163.com}
\and
\IEEEauthorblockN{5\textsuperscript{th} Dongran Han}
\IEEEauthorblockA{\textit{School of Life and Science} \\
\textit{Beijing University of Chinese Medicine}\\
Beijing, China \\
handongr@gmail.com}
\and
\IEEEauthorblockN{6\textsuperscript{th} Zhongzhi Luan}
\IEEEauthorblockA{\textit{School of Computer Science and Engineer} \\
\textit{Beihang University}\\
Beijing, China \\
luan.zhongzhi@buaa.edu.cn}}
\maketitle

\begin{abstract}

Background: Large Language Models (LLMs) have demonstrated a remarkable ability to comprehend and generate natural language in various domains, including the medical field. However, no previous work has studied the performance of LLMs in the context of Traditional Chinese Medicine (TCM), an essential and distinct branch of medical knowledge with a rich history. 

Methods: To bridge this gap, we present a TCM question dataset named TCM-QA, which comprises three question types: single-choice, multiple-choice, and true or false, aimed at examining the LLM’s capacity for knowledge recall and comprehensive reasoning within the TCM domain.The dataset address is https://github.com/yizhen-buaa/TCM-QA-datasets. In our study, we evaluate two settings of the LLM: zero-shot and few-shot settings, while concurrently discussing the differences between English and Chinese prompts. 

Results: Our results indicate that ChatGPT performs best in true or false questions, achieving the highest precision of 0.688, while scoring the lowest precision (0.241) in multiple-choice questions. Furthermore, we observe that Chi- nese prompts outperform English prompts in our evaluations. Additionally, we assess the quality of explanations generated by ChatGPT and their potential contribution to TCM knowledge comprehension. 

Conclusion: By exploring the LLM’s performance in the context of TCM, this paper offers valuable insights into the applicability of LLMs in specialized domains and paves the way for future research in leveraging these powerful models for the advancement of TCM.

\end{abstract}

\begin{IEEEkeywords}
Large Language Models, Prompt Engineering, Traditional Chinese Medicine, Text mining, Artificial Intelligence
\end{IEEEkeywords}

\section{Introduction}
The rapid development of Large Language Models (LLMs) signifies the widespread adoption of artificial intelligence (AI) globally. In addition to its traditional roles in customer support and data management, AI has gained prominence in specialized fields like law\cite{tan2023chatgpt}, medicine\cite{b8}, and content generation\cite{taecharungroj2023can}. Notably, this progress is evident in tasks encompassing intelligent question-answering and content creation. In medicine, LLMs can improve scientific writing, produce formal research articles, and be used as a rapid search engine\cite{b1,b9,b10}. For example, Chat-similar AI models are needed to answer medical questions and to help medical students and doctors engage in medical activities to understand medical knowledge because the practical and applicable model with open education function can improve the independence and autonomy of autonomous learners. It can allow people to acquire knowledge in a way that is easy and free of social pressure\cite{thirunavukarasu2023large}.

ChatGPT, a member of the Generative Pre-training Transformer (GPT) model family, can provide customized and interactive assistance for humans by engaging in complex multi-round conversations\cite{b2}. Based on the GPT-3.5 language model provided by OpenAI, ChatGPT has demonstrated extraordinary abilities in comprehending natural language inputs and formulating relevant responses. One research evaluated the performance of ChatGPT on questions within the scope of the United States Medical Licensing Examination (USMLE) and showed that the model achieves the equivalent of a passing score for a third-year medical student. Additionally, ChatGPT showed remarkable proficiency in providing logic and informational context across the majority of answers\cite{b11}. Furthermore, Microsoft's recent research results show that GPT-4 is significantly better calibrated than GPT-3.5 on USMLE, especially in explaining medical reasoning, personalizing explanations to students, and interactively crafting new counterfactual scenarios around a medical case\cite{b3}. These studies show that, in medicine, the GPT family demonstrates adeptness in engaging in complex conversations, comprehending natural language inputs, and providing contextually relevant responses.

\begin{figure*}[htbp]
\centering
\includegraphics[width=1\linewidth]{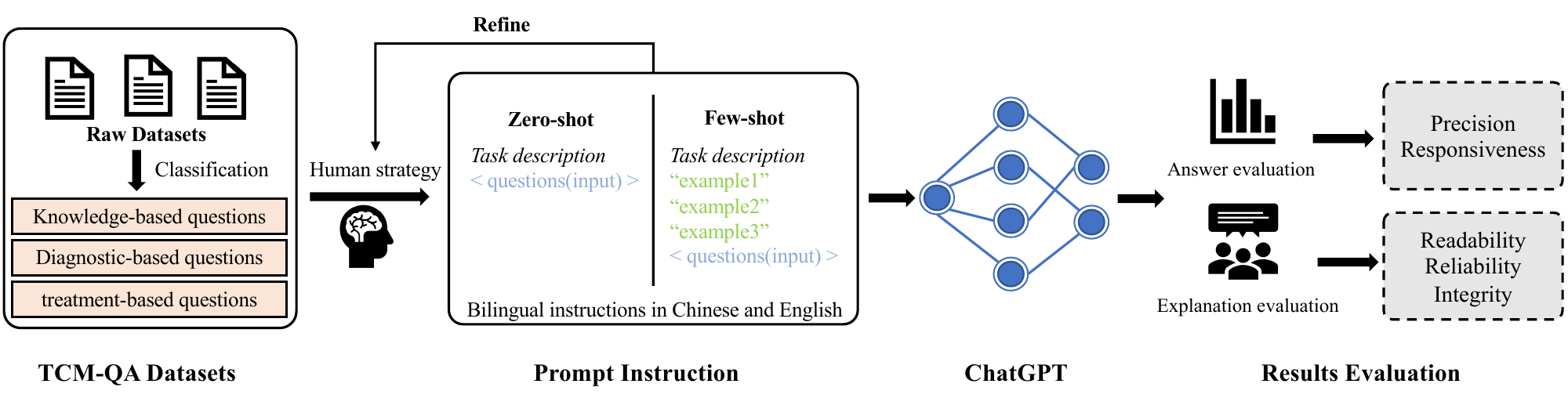}
\caption{The framework of ChatGPT to generate the answers and explanation on TCM knowledge questions}
\label{fig1}
\end{figure*}

“In-context learning” is a novel paradigm in Natural Language Processing (NLP) by employing textual prompts rather than fine-tuning Pre-trained Language Models (PLMs), to reshape downstream tasks\cite{b4}. Unlike traditional supervised learning, prompt-based learning relies on LLMs to directly estimate text probabilities\cite{b5}. This method enables manipulation of model behavior through appropriate prompt selection, facilitating accurate output prediction by pre-trained language models without the necessity of task-specific training. For example, in mental health, if we want to know the emotion of a social media post, we can design a prompt like “post: ‘post’ consider the emotions expressed from this post to answer the question: Is the poster likely to suffer from very severe ‘Condition’? Only return Yes or No, then explain your reasoning step by step”. Then, ChatGPT may be able to fill in the blank with Yes or No and give an explanation\cite{b6}.

Spanning over two millennia in China, Traditional Chinese Medicine (TCM) represents a significant medical legacy. The Chinese have devised a unique treatment framework by integrating human physiology with cosmic and seasonal dynamics. TCM has played a substantial role in managing challenging modern medical conditions such as COVID-19\cite{yang2020traditional}, depression\cite{fu2020kaixinsan}, and cancer care\cite{lu2019randomised}. In TCM, clinical decision-making involves intricate data integration. In addition to standard examinations and tests, TCM practitioners consider the Yin and Yang balance within the body during disease diagnosis and treatment, allowing for a holistic perspective on illnesses\cite{b14}. TCM's disease differentiation approach entails reasoning based on experiential accumulation, complex disease patterns, and syndromes. 

Currently, there is a dearth of research investigating the performance of LLMs in the context of TCM. Therefore, our work is mainly to analyze ChatGPT's understanding ability of TCM knowledge and to evaluate the performance of the ChatGPT in the vertical field of TCM knowledge question-answering. The work is of great significance in exploring the knowledge level of LLM in professional medicine and provides a reference for future researchers to explore the application of LLM in the medical field.

Our study makes three significant contributions:
\begin{itemize}
\item To the best of our knowledge, we construct the first question-answering dataset specifically designed for TCM. We will release TCM-QA dataset for the benefit of the research community. 
\item We assess the performance of ChatGPT in the context of TCM, taking into account both zero-shot and few-shot settings.
\item We examine the quality of the generated explanations by ChatGPT with human evaluation on the question and answer task. To standardize the human evaluation process, we determine three key aspects: readability, reliability, and integrity for assessment. Our results show that ChatGPT maintains outstanding performance in readability and integrity compared with reliability. 
\end{itemize}



\section{Materials and Methods}
The framework of our study performs in Fig.~\ref{fig1}. We first constructed a dataset of TCM problems named TCM-QA and classified it according to the examination focuses (Section \ref{sec.data}). Then, we set up prompt instructions and connected to ChatGPT API, and selected the best-performed prompts by optimizing multiple times (Section \ref{sec.prompts}). Finally, we conducted automatic (Section \ref{sec.auto-eva}) and manual evaluations (Section \ref{sec.human-eva})of the output results.

\subsection{TCM-QA datasets collection}
\label{sec.data}

In this section, we describe our annotation protocol, which consists of three phases. First we automatically extract questions from BaiduWenKu which are expected to be amenable to complex Chinese medicine knowledge. Second, we organized question-answer pairs, eliciting questions which require reasoning. Finally, we classified the TCM-QA datasets to ensure the task.

\textbf{TCM-QA datasets extraction.}
We conducted keyword searches in BaiduWenKu, with a focus on documents containing keywords related to TCM. We particularly prioritized documents with a high proportion of objective questions, as our initial pilot study indicated that objective questions can provide a relatively accurate assessment of large models' performance. We found that the quality of question-and-answer pairs related to TCM fundamental theory, TCM diagnosis, and internal medicine was notably good, and these documents contained a rich set of objective questions. In addition, our dataset also includes a portion of questions related to Chinese Medicine Surgery, Chinese Pediatrics, and Acupuncture.

In total, we retrieved approximately 100 documents from BaiduWenKu. However, we observed a substantial amount of content overlap among these documents. Therefore, we organized and removed duplicate content from these documents.

\textbf{TCM-QA datasets description.}
We corrected the collection of question-answer pairs, with each question having a single standard answer, and our staff is required to locate the exact basis in the reference materials for each answer. To ensure consistency in question-answer pairs, each worker was randomly presented with fifty question-answer pairs and tasked with confirming the answers with reference materials.

Due to the diverse aspects of assessing TCM knowledge in the questions, such as a focus on basic knowledge recall in TCM fundamental theory and a greater emphasis on comprehensive understanding and application in TCM internal medicine, we have designed distinct question categories to align with the different areas of assessment. We provided workers with sample questions in three primary categories for classification purposes, including knowledge reasoning, Symptomatic diagnosis reasoning, and Treatment reasoning questions.

\begin{itemize}
    \item \textit{\textbf{Knowledge Reasoning questions} entail the direct examination of fundamental Chinese medicine knowledge, such as the history of Chinese medicine, prescription efficacy, the five elements' promotion and restriction, and the functions of the five zang-organs.}
\end{itemize}

\begin{itemize}
    \item \textit{\textbf{Symptomatic Diagnosis Reasoning questions} involve a comprehensive analysis of a series of symptom descriptions. These require large models to analyze the symptoms and deduce the disease's diagnostic name or the symptom's name.}
\end{itemize}
\begin{itemize}
    \item \textit{\textbf{Therapeutics Reasoning questions} necessitate the comprehensive analysis of a series of TCM symptom descriptions. Large models are required to select the most suitable treatment plan (prescription) following the diagnosis name or symptom name, as determined through symptom analysis.}
\end{itemize}

Furthermore, to elevate the difficulty level, we introduced multiple-choice and true or false questions. We still required workers to provide supporting evidence for each answer. Additionally, we implemented a review process by randomly assigning questions to two different workers to ensure a bias-free assessment.

We engaged the services of three experts in the field of TCM to complete this annotation task collaboratively. In the end, we gathered a total of 801 question-answer pairs, with each pair undergoing thorough verification.

\begin{figure}[!tb]
\centering
\includegraphics[width=1\linewidth]{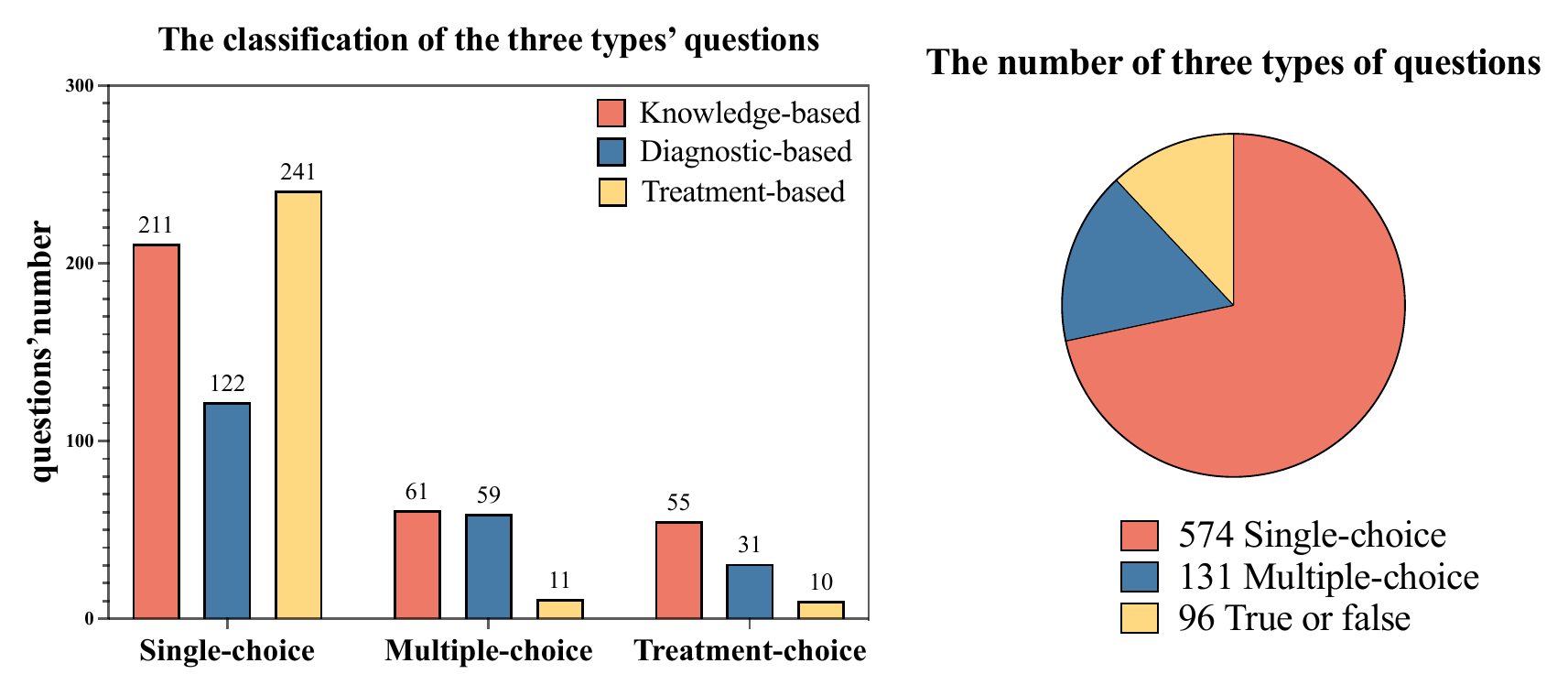}
\caption{The distribution of the three types of questions}
\label{fig2}
\end{figure}

\textbf{TCM-QA datasets analysis.} 
Because there is no publicly available medical knowledge data set in the field of TCM to evaluate the capabilities of ChatGPT on TCM knowledge, we construct a Chinese medicine dataset named TCM-QA, including knowledge and reasoning for testing by referring to the qualification examination of Chinese medicine practitioners and the exercise set of Chinese medicine textbooks (e.g. Basic Theory of Traditional Chinese Medicine\cite{TCM1}, Diagnosis of Traditional Chinese Medicine\cite{TCM2}, and Internal Medicine of Traditional Chinese Medicine\cite{TCM3}). The details on the dataset are provided in Fig.~\ref{fig2}. The dataset contains three types of questions: 574 single-choice questions, 131 multiple-choice questions, and 97 true or false questions. We also divided the TCM questions based on examination focuses, including the reasoning of knowledge, the reasoning of symptom diagnosis, and the reasoning of symptomatic treatment to evaluate the level of ChatGPT's mastery of different knowledge. We will release TCM-QA dataset for the benefit of the research community. https://github.com/yizhen-buaa/TCM-QA-datasets.

\subsection{Prompt instruction}
\label{sec.prompts}

Prompt instructions are tailored to elicit particular information, responses, or behaviors from a language model through targeted input. Multiple iterations of prompt strategies can be utilized to enhance the precision of outcomes in tasks like text generation, question answering, and problem-solving on LLMs\cite{b5}. 

We have developed zero-shot and few-shot prompt settings. Zero-shot settings include \textit{task descriptions} and \textit{target questions}, while few-shot settings involve \textit{TCM Knowledge}, building upon the zero-shot framework.

\textbf{Task descriptions.} The task descriptions include assigning ChatGPT a role-play activity that emphasise ChatGPT's need to think and respond from the perspective of a TCM assistant, requiring ChatGPT to provide step-by-step explanations for its answers. The format of the questions that need to be answered has been briefly introduced.

\textbf{Target questions.} This part allows users to input questions to be answered by ChatGPT. 

\textbf{TCM Knowledge.} This part integrates TCM knowledge and instructing ChatGPT to learn the knowledge prior to answering questions. 
 
Because TCM knowledge and questions are almost expressed in the Chinese language, we also design the prompts in both Chinese and English to evaluate the difference in language thought of ChatGPT. Our study investigated whether different prompting strategies improve ChatGPT's ability to answer TCM questions. We present illustrative prompts for both zero-shot and few-shot settings as follows:




\begin{tcolorbox}[enhanced,attach boxed title to top left={yshift=-3mm,yshifttext=-1mm},  
  colback=green!5!white,colframe=green!50!black,colbacktitle=green!60!black,  
  title=Zero-shot Prompting,fonttitle=\bfseries,  
  boxed title style={size=small,colframe=green!30!black} ]  
  Your task is to select one answer from the following single-choice questions and answers about traditional Chinese medicine knowledge. Give your answer as one of `A', `B', `C', `D', and `E' from the following answers. Please choose one answer and give your explanation by using at most 1000 words in Chinese. 
  \\ \\
  **Single-choice questions: ** ```\{What is the therapeutic principle of phlegm retention?
A. Diffusing the lung B. Tonifying the spleen C. Warmly resolving the phlegm D. Tonifying the kidney E. Sweating\}'''
    \\ \\
  **Answer: **
\end{tcolorbox}  

\begin{tcolorbox}[enhanced,attach boxed title to top left={yshift=-3mm,yshifttext=-1mm},  
  colback=blue!5!white,colframe=blue!40!black,colbacktitle=blue!60!black,  
  title=Few-shot Prompting,fonttitle=\bfseries,  
  boxed title style={size=small,colframe=blue!20!black} ]  
  There are some examples of statements (with answers) about traditional Chinese medicine Knowledge. \\ \\
  ```few-shot-example 1''', \\
  ```few-shot-example 2''', \\
  ```few-shot-example 3''', \\ \\
    Your task is to determine whether the following statements about traditional Chinese medicine knowledge are correct. Give your answer as either ```Yes''' or ``No'''. Please choose the answer and give your explanation by using at most 1000 words in Chinese.
  \\ \\
    **Statements: ** ```\{The heart controls blood circulation. And the liver stores blood. Clinically, heart and liver blood deficiency can appear at the same time.\}'''

    **Answer: **”
\end{tcolorbox}

\subsection{Automatic evaluation for objective results}
\label{sec.auto-eva}
We use precision as the first evaluation metric. For single-choice and true or false questions, Correct answers are scored 1 point, and incorrect answers are scored 0 points. For multiple-choice questions, 1 point will be scored for complete, correct answers, 0.5 points for partially correct answers, and 0 points for all wrong options. The total score of all questions is counted as Correct Numbers (CN). The total number of questions is counted as Total Numbers (TN).

The precision is defined in Equation 1:
\begin{equation}
\text{Precision (P)}=\frac{CN}{TN}\label{eq}
\end{equation}

We take the responsiveness as the second metric. In some cases, ChatGPT didn't clearly answer a certain question but first explained the content. For such results, we consider them to be “unresponsive.” The “Precision in Responsive” in this situation means the Correct Numbers (CN) in the Responsive Numbers (RN) provided by ChatGPT. The responsive is defined in Equation 2:
\begin{equation}
\text{Precision in Responsive (PR)} =\frac{CN}{RN}\label{eq}
\end{equation}

\subsection{Human evaluation for explanation ability}
\label{sec.human-eva}
We examine the quality of the generated explanations with a human evaluation of the choice and true and false questions tasks. To evaluate the performance of ChatGPT, we order ChatGPT to generate explanations within 1000 words (in Chinese) for the questions with the same prompts. We separately select 150 results generated by ChatGPT that answer correct and wrong to enable fair comparisons of their explanations, including knowledge-based, diagnostic-based, and treatment-based questions. 30 more no-responses answers are also collected for error analysis. 

To standardize the human evaluation process, we determine three critical aspects for assessment:
\begin{itemize}
\item \textbf{Readability}. It includes the coherence and the fluency of the explanations. The evaluators are supposed to assess if generated explanation is well-structured, easy to read and understand and free of grammatical mistake or syntax error:

\begin{enumerate}
    \item Unordered, incoherent, difficult to read or understand and remaining numerous errors;
    \item Poorly structured, mostly coherent, easy to read and understand and including several errors;
    \item Well-structured, fully coherent, completely fluent, intelligible and error-free.
\end{enumerate}

\item \textbf{Reliability}. It indicates the comprehension of the questions and the trustworthiness of the explanations to support the answer results. The evaluators are able to assess whether the description and the explanation are based on facts, contain reliable information and perform great reasoning according to the given information:

\begin{enumerate}
    \item Unreliable or inconsistent information, poor comprehension and wrong reasoning;

    \item Mostly reliable information with some inconsistencies and appropriate comprehension and reasoning;

    \item Completely reliable information, inconsistency-free, true comprehension and great reasoning.
\end{enumerate}

\item \textbf{Integrity}. It measures how well the explanations comprise all relevant knowledge. The evaluators should assess if the explanation provides sufficient context and full description:

\begin{enumerate}
    \item Omits significant information from relevant knowledge;

    \item Partially complete with minor omissions;

    \item Completely comprising relevant knowledge.
\end{enumerate}
\end{itemize}

Based on the above definitions, the expert further determines the assessment criteria, where each aspect is divided into three standards rating from 1 to 3. Higher ratings reflect more satisfactory performance in the corresponding aspect, and three denotes approaching human performance. Each explanation is assigned a score by three annotators with TCM background for each aspect, and summarize the results by the expert.

We further evaluate the quality of the annotations by calculating the inter-evaluator agreement: Fleiss’ Kappa statistics for each aspect. Any annotations with a majority vote are considered as reaching an agreement.
\subsection{Language model}

We evaluate the latest OpenAI's GPT-3.5 model~\cite{ouyang-etal-2022-instructgpt} (`\texttt{gpt-3.5-turbo}', v. March 2023), the first available ChatGPT model. We uses the \texttt{ChatCompletion} API\footnote{https://platform.openai.com/docs/guides/chat} providing one instruction-following example at a time as input with a generation limit up to 1000 tokens.

\section{Results}
\subsection{The overall performance}
Table~\ref{tab: zero-shot-results} and Table~\ref{tab: few-shot-results} presented the TCM knowledge performance of ChatGPT on zero-shot and few-shot settings. The results show that ChatGPT presented the best performance in true or false questions and presented the lowest score in multiple questions, indicating that ChatGPT is better at reasoning simple binary conditions, while the results obtained correct results in more complex conditions are limited. 

From the perspective of overall precision (P) and precision in responsive (PR), the impact of different prompts on the answer given by ChatGPT is not significant, indicating that simply setting different prompt words to improve ChatGPT's understanding of TCM topics is not entirely effective. The PR score of true or false questions and multiple questions is higher than single-choice questions. For the effective response of the model, it is easier to make judgments on true or false questions than on choice questions. 

\begin{table}[!htbp]
\centering
\caption{Zero-shot performance of ChatGPT on TCM-QA}
\label{tab: zero-shot-results}
\begin{tabular}{c|lcc} 
\toprule
\multicolumn{1}{l}{\textbf{Types of questions}}                                                & \textbf{Metric} & \textbf{English}    & \textbf{Chinese}     \\ 
\hline
\multirow{2}{*}{\begin{tabular}[c]{@{}c@{}}single-choice\\(total=574)\end{tabular}}   & P    & 0.314 & 0.334  \\
                                                                                      & PR      & 0.340 & 0.36   \\
                                                                                   
\hline\hline
\multirow{2}{*}{\begin{tabular}[c]{@{}c@{}}multiple-choice\\(total=131)\end{tabular}} & P    & 0.187 & 0.240  \\
                                                                                      & PR      & 0.196 & 0.242  \\
                                                                              
\hline\hline
\multirow{2}{*}{\begin{tabular}[c]{@{}c@{}}true or false\\(total=97)\end{tabular}}    & P   & 0.552 & 0.625  \\
                                                                                      & PR      & 0.552 & 0.625  \\
                                                                                    
\bottomrule
\end{tabular}
\end{table}

\begin{table}[!htbp]
\centering
\caption{Few-shot performance of ChatGPT on TCM-QA}
\label{tab: few-shot-results}
\begin{tabular}{c|lll} 
\toprule
\multicolumn{1}{l}{\textbf{Types of questions}}                                                & \textbf{Metric} & \textbf{English}    & \textbf{Chinese}     \\ 
\hline
\multirow{2}{*}{\begin{tabular}[c]{@{}c@{}}single-choice\\(total=574)\end{tabular}}   & P~     & 0.258 & 0.328  \\
                                                                                      & PR      & 0.330 & 0.346  \\
                                                                                    
\hline\hline
\multirow{2}{*}{\begin{tabular}[c]{@{}c@{}}multiple-choice\\(total=131)\end{tabular}} & P~     & 0.210 & 0.225  \\
                                                                                      & PR      & 0.241 & 0.229  \\
                                                                                     
\hline\hline
\multirow{2}{*}{\begin{tabular}[c]{@{}c@{}}true or false\\(total=97)\end{tabular}}    & P~     & 0.583 & 0.788  \\
                                                                                      & PR      & 0.583 & 0.688  \\
                                                                              
\bottomrule
\end{tabular}
\end{table}

\subsection{The Fine-Grained Results}
In this section, we present a detailed analysis of our findings by examining the performance on different types of questions, namely single-choice, multiple-choice, and true or false questions. By analyzing the results in a fine-grained manner, we aim to provide insights into the strengths and weaknesses of ChatGPT and to understand its implications on different question types. 

\subsubsection{Single-choice questions}

The experimental results on single-choice questions are presented in Table~\ref{Table2}. 

\begin{table}[]
\centering
\caption{The precision of correct answers on single-choice questions}
\begin{tabular}{ccccc}
\toprule
                                                                              & \textbf{\begin{tabular}[c]{@{}c@{}}zero-shot \\ (English)\end{tabular}} & \textbf{\begin{tabular}[c]{@{}c@{}}zero-shot\\ (Chinese)\end{tabular}} & \textbf{\begin{tabular}[c]{@{}c@{}}few-shot\\  (English)\end{tabular}} & \textbf{\begin{tabular}[c]{@{}c@{}}few-shot\\  (Chinese)\end{tabular}} \\ \hline
\textbf{\begin{tabular}[c]{@{}c@{}}Knowledge-based\\ total=211\end{tabular}}  & 0.332                                                                   & 0.313                                                                  & 0.332                                                                  & 0.327                                                                  \\
\textbf{\begin{tabular}[c]{@{}c@{}}Diagnostic-based\\ total=122\end{tabular}} & 0.385                                                                   & 0.402                                                                  & 0.295                                                                  & 0.369                                                                  \\
\textbf{\begin{tabular}[c]{@{}c@{}}Treatment-based\\ total=241\end{tabular}}  & 0.261                                                                   & 0.320                                                                  & 0.174                                                                  & 0.307                                                                  \\ \bottomrule
\end{tabular}
\label{Table2}
\end{table}

\begin{figure}[htbp]
\centering
\includegraphics[width=1\linewidth]{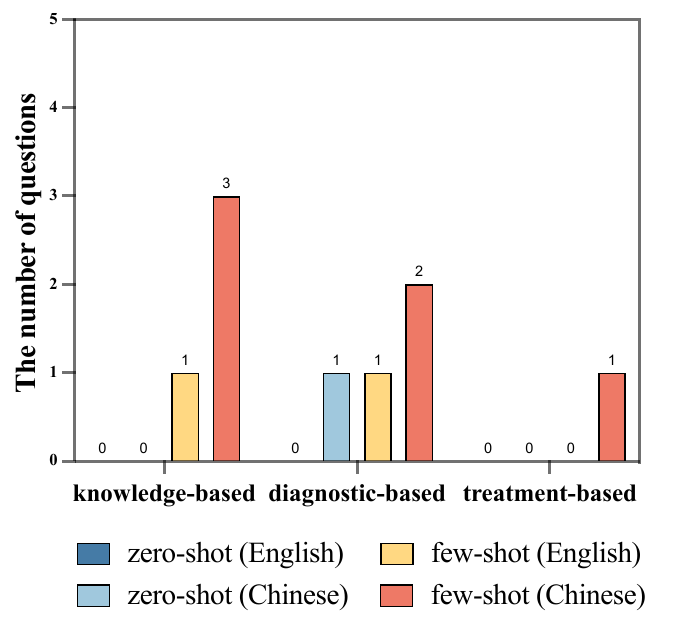}
\caption{The unresponsive results of ChatGPT on single choice questions.}
\label{fig3}
\end{figure}

ChatGPT scored the best performance on the TCM diagnostic-based questions for the zero-shot prompt with 0.402 (Chinese zero-shot). ChatGPT scored the best performance on the TCM diagnostic-based questions for the few-shot prompt with 0.369 (Chinese few-shot). Except for knowledge-based questions, ChatGPT achieves higher precision by using Chinese language in the prompts. Fig.~\ref{fig3} reflects the unresponsive results of ChatGPT on different content single-choice questions. No response occurred most often for knowledge-based and diagnostic-based questions, especially those running the prompt with few examples.

\subsubsection{Multiple-choice questions}

The experimental results on multiple choice questions are presented in Table~\ref{Table3}. ChatGPT scored the best performance on the TCM treatment-based questions for the zero-shot prompt with 0.273 (Chinese zero-shot). For the few-shot prompt, ChatGPT scored the best performance on the TCM knowledge-based questions with 0.230 (Chinese few-shot). ChatGPT obtained higher precision by using Chinses language in the prompts. All multiple questions are responded by ChatGPT. 

\begin{table}[]
\centering
\caption{The precision of correct answers on multiple-choice questions}
\begin{tabular}{ccccc}
\toprule
                                                                             & \textbf{\begin{tabular}[c]{@{}c@{}}zero-shot \\ (English)\end{tabular}} & \textbf{\begin{tabular}[c]{@{}c@{}}zero-shot\\ (Chinese)\end{tabular}} & \textbf{\begin{tabular}[c]{@{}c@{}}few-shot\\  (English)\end{tabular}} & \textbf{\begin{tabular}[c]{@{}c@{}}few-shot\\  (Chinese)\end{tabular}} \\ \hline
\textbf{\begin{tabular}[c]{@{}c@{}}Knowledge-based\\ total=61\end{tabular}}  & 0.148                                                                   & 0.230                                                                  & 0.213                                                                  & 0.230                                                                  \\
\textbf{\begin{tabular}[c]{@{}c@{}}Diagnostic-based\\ total=59\end{tabular}} & 0.229                                                                   & 0.246                                                                  & 0.229                                                                  & 0.229                                                                  \\
\textbf{\begin{tabular}[c]{@{}c@{}}Treatment-based\\ total=11\end{tabular}}  & 0.182                                                                   & 0.273                                                                  & 0.091                                                                  & 0.182                                                                  \\ \bottomrule
\end{tabular}
\label{Table3}
\end{table}

\subsubsection{True or false questions}


The experimental results on true or false questions are presented in Table~\ref{Table4}. For the zero-shot prompt, ChatGPT scored the best performance on the TCM treatment-based questions with 0.70 (Chinese zero-shot). For the few-shot prompt, ChatGPT scored the best performance on the TCM knowledge-based questions with 0.745 (Chinese few-shot). For true or false questions, ChatGPT obtained higher precision by using Chinses language prompts, and a few examples could improve the precision of the answers. Only one question occurred unresponsive in the zero-shot prompt (Chinese) with knowledge-based on true or false questions.

\begin{table}[]
\centering
\caption{The precision of correct answers on true or false questions}
\begin{tabular}{ccccc}
\toprule
                                                                             & \textbf{\begin{tabular}[c]{@{}c@{}}zero-shot \\ (English)\end{tabular}} & \textbf{\begin{tabular}[c]{@{}c@{}}zero-shot\\ (Chinese)\end{tabular}} & \textbf{\begin{tabular}[c]{@{}c@{}}few-shot\\  (English)\end{tabular}} & \textbf{\begin{tabular}[c]{@{}c@{}}few-shot\\  (Chinese)\end{tabular}} \\ \hline
\textbf{\begin{tabular}[c]{@{}c@{}}Knowledge-based\\ total=55\end{tabular}}  & 0.600                                                                   & 0.673                                                                  & 0.636                                                                  & 0.745                                                                  \\
\textbf{\begin{tabular}[c]{@{}c@{}}Diagnostic-based\\ total=31\end{tabular}} & 0.484                                                                   & 0.516                                                                  & 0.484                                                                  & 0.581                                                                  \\
\textbf{\begin{tabular}[c]{@{}c@{}}Treatment-based\\ total=10\end{tabular}}  & 0.500                                                                   & 0.700                                                                  & 0.600                                                                  & 0.700                                                                  \\ \bottomrule
\end{tabular}
\label{Table4}
\end{table}

\subsection{The results of explanation ability}

The above subsection shows that Chinese zero-shot and few-shot prompts can enhance ChatGPT’s performance in the TCM-QA dataset, especially in multiple-choice and true or false questions. Moreover, we can also ask ChatGPT to explain its step-by-step reasoning for each answer by prompt instruction. This process can significantly improve the explainability of AI algorithm, which is a crucial advantage compared with most previous black-box methods. In this subsection, we designed a human evaluation strategy to gain a clear view of ChatGPT’s explainability ability. We provided an example of ChatGPT's explanation ability on multiple choice questions.

\begin{tcolorbox}[enhanced,attach boxed title to top left={yshift=-3mm,yshifttext=-1mm},  
  colback=yellow!5!white,colframe=yellow!50!black,colbacktitle=yellow!60!black,  
  title= An example of explanation,fonttitle=\bfseries,  
  boxed title style={size=small,colframe=yellow!30!black} ]  
  Question: 
  
  The main physiological function of the gallbladder is: A.synthesizing blood; B.excreting waste products; C.secreting bile; D.storing bile; E.excreting bile
  
  Explanation: 
  
  The correct options for this question are: C and D.
  
  Liver produces bile, which is stored in the gallbladder. The gallbladder releases the bile into the small intestine to aid in the digestion and absorption of fats. Therefore, the main physiological function of the gallbladder is to store and release bile. Therefore, option C (secreting bile) and option D (storing bile) are the correct answers.
  
  A (synthesizing blood) is a function of the spleen in traditional Chinese medicine (TCM), not the gallbladder. B (excreting waste products) and option E (excreting bile) are not the primary functions of the gallbladder. While the gallbladder does excrete bile, this is not its primary function and is not the reason for its existence. 
  
\end{tcolorbox}  

\begin{figure*}[htbp]
\centering
\includegraphics[width=1\linewidth]{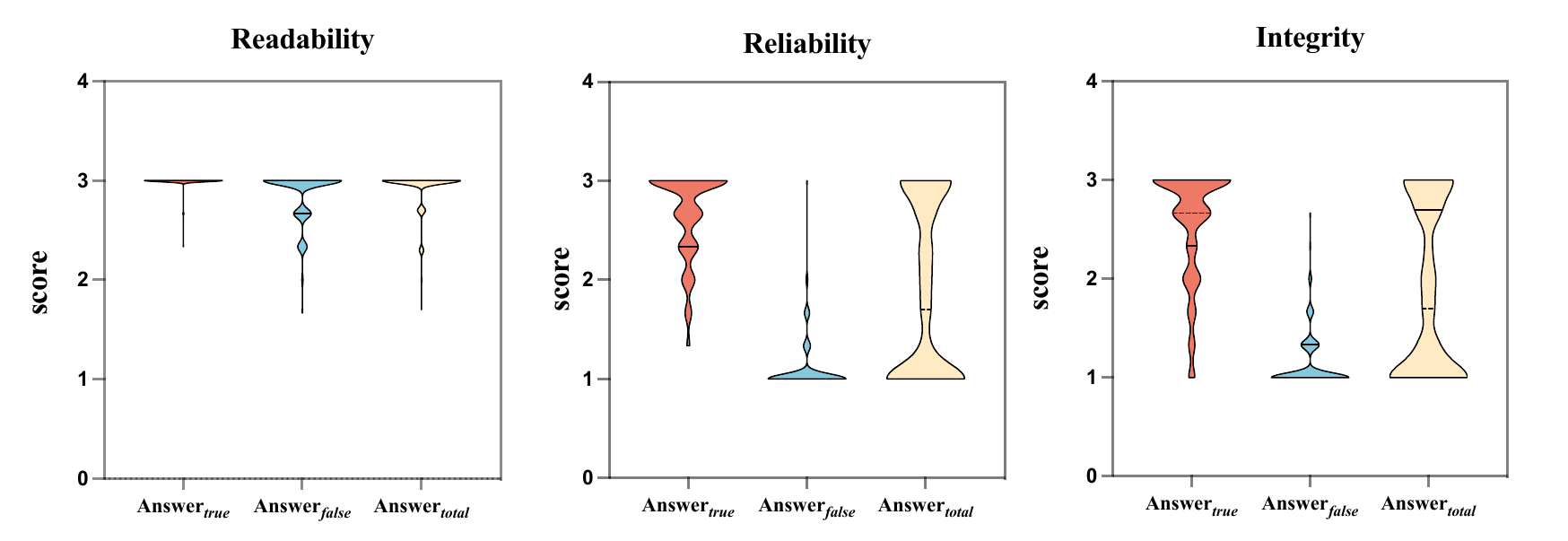}
\caption{Violin plots of the aggregated human evaluation scores for each aspect. Solid lines denote the quartile scores, and dotted lines denote the median scores.}
\label{fig5}
\end{figure*}





\begin{table}[]
\centering
\caption{Fleiss’ Kappa of human evaluations on ChatGPT explanation results.}
\begin{tabular}{llll} 
\toprule
ChatGPT's results & Rea.  & Rel.  & Int.  \\ \hline
Answer true    & 0.244 & 0.37  & 0.432 \\
Answer false      & 0.244 & 0.456 & 0.213 \\
Answer total             & 0.247 & 0.646 & 0.594 \\ \bottomrule
\end{tabular}
\label{Table kappa}
\end{table}

\begin{table}[]
\centering
\caption{Average score statistics of human evaluations on ChatGPT explanation results.}
\begin{tabular}{ccccc}
\toprule
                  &             & \multicolumn{3}{c}{Average score} \\ \hline
ChatGPT's results & sample num. & Rea.      & Rel.      & Int.      \\ \hline
Answer true    & 160         & 2.991     & 2.67      & 2.571      \\
Answer false      & 160         & 2.867     & 1.094     & 1.142     \\
Total             & 320         & 2.929     & 1.882     & 1.856     \\ \bottomrule
\end{tabular}
\label{Table average}
\end{table}




Firstly, the three annotators reach an agreement in most cases of evaluation. The Fleiss’ Kappa statistics results are presented in Table~\ref{Table kappa}. All Fleiss’ Kappa statistics achieve at least fair agreement($\geq$0.21). The reliability statistics reach a moderate agreement($\geq$0.61), and integrity statistics reach a substantial agreement ($\geq$0.41). These outcomes prove the quality of the human evaluation results. We also summarized the average scores for each scenario in Table~\ref{Table average}. We aggregate the median and quartile for three scenarios of ChatGPT by the violin plot: the true answer, false answer, and overall answer, and the distributions are presented in Fig.~\ref{fig5}. 

According to the violin plot of the aggregated scores, Answer$_{true}$ almost achieves an average score of 3.0 in readability and maintains outstanding performance, which shows that ChatGPT can achieve human-level responses regarding fluency and coherence. In reliability, Answer$_{true}$ achieves a median score of 2.4 and 2.6 scores on average, showing ChatGPT as a very trustworthy reasoner in supporting its classifications. In integrity, Answer$_{true}$ obtains over 2.5 scores on average, indicating that ChatGPT can cover most of the relevant content in the questions to explain its classifications. Answer$_{true}$ has an average score of over 2.5, proving that ChatGPT can generate human-level explanations for correct classifications regarding readability, reliability, and integrity. On the other hand, Answer$_{false}$ achieves 3.0 scores on readability but low scores on reliability and integrity, proving that ChatGPT can generate illusions when exceeding its' knowledge range.

\section{Discussion}
In recent years, LLMs have exhibited remarkable proficiency across various domains. General LLMs have achieved notable results through extensive data utilization, substantial computational resources, and self-supervised pre-training without manual annotations. Further enhancing these general LLMs via fine-tuning with limited additional data holds substantial promise for significant advancements in downstream specialized domains.\cite{b15}.

ChatGPT, featuring an expansive parameter count of 175 billion, enhances its performance by autonomously learning from web text data and continuously updating its weights\cite{b16}. It possesses formidable cognitive capabilities, encompassing natural language comprehension and reasoning skills. Notably, OpenAI researchers have improved the model's conversational prowess through "prompt engineering," a technique involving step-by-step guiding prompts. Preceding this study, ChatGPT demonstrated commendable performance across specialized domains, including medicine\cite{b17}, business administration\cite{b18,b19}, and law\cite{b20}, achieving noteworthy scores without domain-specific fine-tuning.

Model training in the biomedical domain has been historically hampered by the challenges of acquiring specialized medical data, resulting in high costs and inefficiencies\cite{b21,b22,b23}. The conventional approach necessitates extensive labeled biomedical data for training models to comprehend biomedical text and engage in knowledge-based reasoning. The emergence of LLMs in AI, however, presents the opportunity to directly apply generic models in the biomedical field, alleviating the need for medical data fine-tuning. The notable performance of ChatGPT and GPT-4 in the USMLE exam underscores the considerable potential of generic models within the medical domain.

This study investigates the applicability of a new paradigm to enhance Question and Answer understanding in TCM. Through our designed experiments and comprehensive analysis of ChatGPT's TCM comprehension, we find a moderate level of performance. Our assessment covers multiple aspects, including ChatGPT's medical knowledge recall and holistic reasoning abilities. We employ three question types: knowledge-based (retrieving medical knowledge), diagnostic (explaining diseases or syndromes from symptoms), and treatment-related (analyzing patient scenarios for treatment inferences). 

\begin{figure}[!tb]
\centering
\includegraphics[width=0.65\linewidth]{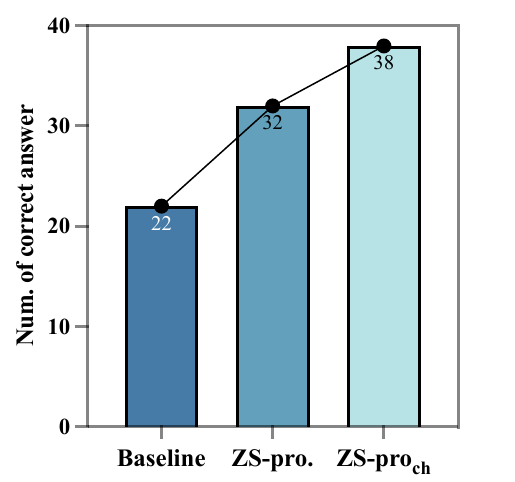}
\caption{The prompting instruction refine results on one hundred single choice questions.}
\label{fig6}
\end{figure}

We designed three pre-experiments to prompt instruction. As a starting point, we requested the output result of ChatGPT's dialogue box(\textit{Baseline}). Next, we created a zero-shot prompt (\textit{ZS-pro}) that lists the format of the target questions and requires ChatGPT to act as a TCM practitioner and generate explanations. Finally, we translated the prompts into Chinese and examined the performance of TCM knowledge across different languages. According to the results presented in Figure~\ref{fig6}, a prompt setting and the use of the Chinese language were found to improve ChatGPT's performance in understanding TCM knowledge, which may be due to the fact that the questions are written in Chinese, and the Chinese prompt can express the instruction more precisely. 

Results reveal ChatGPT 's proficiency and show better performance in comprehending knowledge-based and diagnostic questions. The complexity of comprehensive treatment reasoning, requiring nuanced TCM knowledge, likely accounts for the disparity. ChatGPT's current understanding of TCM seems inadequate for this level of proficiency.

By analyzing the results, we find ChatGPT still has a relatively shallow understanding of TCM knowledge, which may be related to the training data it received. Over 50 percent of ChatGPT's GPT-3.5 pre-trained models are in English, with the primary dataset, WebText2, originating from English-speaking countries such as the United States, United Kingdom, Canada, and Australia\cite{b24}. The limited exposure to knowledge related to TCM by ChatGPT is primarily due to the majority of its training data being in English. Furthermore, the fact that our questions are presented in Chinese contributes to this limitation. The prompt language in which the questions are posed is consistent with ChatGPT's training data, which may contribute to improved performance in responding to questions in Chinese. It explains why prompts expressed in Chinese have yielded higher precision. Essentially, the model's performance in TCM is constrained by its linguistic training bias, which affects its ability to fully comprehend and respond to TCM-related questions.

LLMs like ChatGPT can enhance interpretability by setting explicit prompts, enabling step-by-step explanations, and can judge the rationality by humans—a significant advantage over black-box models. Our study evaluates the explanations generated by ChatGPT. We limited ChatGPT's explanations to 1000 words per question to meet response time and field constraints. The results show the average number of words in the ChatGPT's explanations is 207. We observed instances where ChatGPT meticulously analyzed particular keywords in the question without comprehending the overall sentence context, potentially leading to incorrect answers. Moreover, ChatGPT frequently generated "illusions" by inventing erroneous TCM knowledge to substantiate its rationale. These observations underscore ChatGPT's insufficient TCM knowledge reservoir for addressing professional queries.

\section{Limitation}
Although we have devised zero-shot and few-shot prompts for each question type, the results do not yet provide substantial evidence of significant differences resulting from varied prompt settings. Prior research on hint words in ChatGPT has demonstrated their potential to enhance problem comprehension and interpretable answer generation. More advanced prompt techniques, including thought chain of thought prompting\cite{b25,b26}, self-consistency prompting\cite{b27}, and retrieval tools for accessing information\cite{b28}, have significantly improved model performance. Consequently, exploring diverse prompting strategies holds promise for deepening ChatGPT's comprehension of TCM knowledge.

This study has focused on choice and true or false questions encompassing TCM knowledge, which constitute a substantial but incomplete portion of the Chinese medicine practitioner qualification examination and TCM textbook content. Specifically, the examination stages of TCM knowledge also involves practical interaction and patient case analysis. These necessitate examinees to engage with patients in a simulated setting, perform comprehensive analyses, and articulate patient scenarios. This aspect was not evaluated in ChatGPT. Furthermore, the distribution of multiple-choice and true-false questions is unbalanced, with limited questions focused on treatment-based inquiries. This imbalance might restrict the evaluation of ChatGPT's comprehensive reasoning abilities.

Due to cost constraints, our evaluation was restricted to the performance of a single representative LLM, ChatGPT (GPT-3.5-turbo). Given the swift evolution of LLMs, numerous other notable models exist, such as GPT-4, chatGLM, and LLaMa2. A more exhaustive analysis of LLM capabilities in TCM knowledge is warranted.

In summary, diversifying datasets, employing advanced prompt word strategies, and broadening evaluations across multiple models could facilitate a more comprehensive assessment of the performance of LLMs in TCM. These approaches would also enable a multi-dimensional exploration of the development potential of large models within TCM scenarios.
\section{Conclusion}
In this work, we comprehensively investigated the effects of Chat-GPT on TCM knowledge retention and reasoning ability and between different prompting strategies. We focused on questions related to TCM knowledge, including 574 single-choice questions, 131 multiple-choice questions, and 96 true or false questions. We also explored ChatGPT's potential to answer different types of TCM-related questions, such as knowledge-based, diagnostic, and treatment-based. We used precision and responsiveness as evaluation metrics and performed a rigorous human evaluation to assess the quality of the explanations generated by ChatGPT. The findings reveal heightened precision in binary questions like true or false, while more intricate queries yield suboptimal results. We also demonstrate the powerful interpretability of ChatGPT, particularly in fluently generating professional language pertinent to TCM.  Moreover, our analysis of ChatGPT's explanation results showed that LLMs can sometimes produce reliable explanations for their decisions. However, this reliability hinges on their training data, occasionally resulting in false "illusions."

In conclusion, tasks such as reasoning about clinical problems and complex knowledge recall are still challenging for ChatGPT, and our attempts to incorporate appropriate prompting did not yield superior results. We posit that enhancing ChatGPT's performance in knowledge and reasoning within specialized domains is feasible through instruction fine-tuning and the integration of high-quality data.


\section*{Author Contribution Statement}
YL collected and analyzed the TCM-QA datasets, and was a major contributor in writing the manuscript. LQ  was involved in verifying the data results and polishing the language. SH and XJ coded the ChatGPT's prompt words. DH and ZL participated in the revision of the manuscript. All authors read and approved the final manuscript.

\bibliographystyle{IEEEtran}
\bibliography{Mybib}

\end{document}